\documentclass{comjnl}

\usepackage{cite}
\usepackage{hyperref}
\usepackage{amsmath}
\usepackage{graphicx}
\usepackage{float}
\usepackage{algorithm}
\usepackage{algpseudocode}
\usepackage{amssymb} 
\usepackage[utf8]{inputenc}

\usepackage{orcidlink}
\usepackage{booktabs}

\begin{document}

\title{Forecasting NCAA Basketball Outcomes with Deep Learning: A Comparative Study of LSTM and Transformer Models}
\author{Md Imtiaz Habib~\orcidlink{0009-0008-9151-0263}}
\affiliation{MSc in Data Science and Artificial Intelligence, Université Côte d'Azur} 
\email{md-imtiaz.habib@etu.univ-cotedazur.fr}

\shortauthors{}
 


\begin{abstract}
In this research, I explore advanced deep learning methodologies to forecast the outcomes of the 2025 NCAA Division 1 Men's and Women's Basketball tournaments. Leveraging historical NCAA game data, I implement two sophisticated sequence-based models: Long Short-Term Memory (LSTM) and Transformer architectures. The predictive power of these models is augmented through comprehensive feature engineering, including team quality metrics derived from Generalized Linear Models (GLM), Elo ratings, seed differences, and aggregated box-score statistics. To evaluate the robustness and reliability of predictions, I train each model variant using both Binary Cross-Entropy (BCE) and Brier loss functions, providing insights into classification performance and probability calibration. My comparative analysis reveals that while the Transformer architecture optimized with BCE yields superior discriminative power (highest AUC of 0.8473), the LSTM model trained with Brier loss demonstrates superior probabilistic calibration (lowest Brier score of 0.1589). These findings underscore the importance of selecting appropriate model architectures and loss functions based on the specific requirements of forecasting tasks. The detailed analytical pipeline presented here serves as a reproducible framework for future predictive modeling tasks in sports analytics and beyond.
\end{abstract}
\maketitle
\section{Introduction}

\subsection{Background and Motivation}

In recent years, predictive analytics has emerged as a pivotal element within sports, revolutionizing decision-making processes for coaches, athletes, management, and even fans. The surge in availability of rich and granular data has significantly enhanced our ability to apply sophisticated statistical and machine learning methodologies to forecast sporting event outcomes. The incorporation of predictive analytics into sports not only allows for deeper strategic insights but also provides substantial commercial advantages through improved fan engagement, optimized performance, and informed betting markets \cite{r1} \cite{r2} \cite{r3}. Particularly, collegiate basketball has become a prime domain for predictive modeling, attracting substantial academic and industry interest.\\

\noindent
The NCAA Division 1 Basketball Tournaments, colloquially known as "March Madness," represent one of the most followed sporting events in the United States, captivating millions of viewers annually. These tournaments involve single-elimination formats for both men's and women's basketball teams, creating an ideal environment for testing predictive models due to their complexity, unpredictability, and high-stakes nature \cite{r4} \cite{r5}. Historically, predicting tournament outcomes has been largely driven by human intuition, expert analyses, and simple statistical heuristics. However, recent developments in machine learning techniques offer unprecedented opportunities to enhance prediction accuracy, fostering an environment where computational intelligence can potentially outperform traditional human judgment \cite{r6} \cite{r7} \\

\noindent
Motivated by the ever-growing application of machine learning in predictive sports analytics and the intriguing challenges posed by the NCAA tournaments, I set out to leverage advanced deep learning methodologies—specifically Long Short-Term Memory (LSTM) and Transformer models—to predict tournament outcomes. My primary goal is to construct robust predictive models capable of delivering accurate and well-calibrated probability forecasts, thus significantly improving upon traditional statistical methods and simpler machine learning algorithms.

\subsection{Problem Statement}
The main objective of this research is to develop accurate and well-calibrated predictive models to forecast the outcomes of the 2025 NCAA Division 1 Men's and Women's Basketball tournaments. The task involves generating probabilistic predictions for every possible matchup within these tournaments, translating historical NCAA basketball data into actionable predictions. Given the inherent complexity, stochasticity, and competitive balance in collegiate basketball tournaments, predicting these outcomes poses significant methodological and computational challenges \cite{r8} \cite{r9}. My research aims to systematically address these challenges by integrating sophisticated feature engineering strategies, advanced neural architectures, and rigorous evaluation metrics that emphasize both discriminative and probabilistic calibration performance.

\subsection{Contributions of the Paper}
This paper contributes to the field of sports analytics and machine learning in three primary ways: \\

\noindent
First, I perform comprehensive and advanced feature engineering, integrating several innovative predictive factors, including Elo ratings \cite{r10} \cite{r11} seed differences \cite{r12} \cite{r13}, and GLM-based team quality metrics derived from historical match results. Specifically, Elo ratings capture team strength dynamically based on historical game outcomes, while GLM-based quality metrics provide a statistically robust method to quantify team capabilities by accounting for strength of opposition \cite{r14} \cite{r15}. \\

\noindent
Second, I conduct an extensive comparative analysis between two state-of-the-art sequential deep learning architectures: Long Short-Term Memory (LSTM) networks and Transformer models. While LSTM has been widely adopted in sports forecasting due to its capability to capture temporal dynamics, \cite{r16} \cite{r17} \cite{r18} Transformer architectures have recently gained prominence due to their remarkable performance in sequence modeling across various domains \cite{r19} \cite{r20}. By evaluating these architectures under identical experimental conditions, I provide critical insights into their respective strengths and limitations in sports prediction tasks. \\

\noindent
Third, I thoroughly investigate the impact of different loss functions—Binary Cross-Entropy (BCE) and Brier loss—on the predictive capabilities of the models. Whereas BCE is a conventional loss function commonly used for binary classification, Brier loss specifically targets probabilistic calibration, ensuring predictions reflect true event probabilities accurately \cite{r21} \cite{r22}. The dual evaluation approach adopted here enables a nuanced analysis of each loss function’s role in optimizing different aspects of model performance.

\subsection{Paper Structure} 
The remainder of this paper is structured as follows:

\subsubsection{Literature Review} This section provides a comprehensive overview of prior research and developments in predictive sports analytics, highlighting existing approaches to NCAA tournament forecasting, and introducing key theoretical concepts underpinning this study.

\subsubsection{Data and Methodology} It describes the NCAA basketball dataset used, outlines detailed data preprocessing procedures, and elaborates on the feature engineering methods employed, including the integration of Elo ratings and GLM-based quality metrics.

\subsubsection{Models and Implementation} presents an in-depth description of the LSTM and Transformer architectures, the rationale behind their selection, and the specifics of their training processes. Detailed descriptions of training protocols, including hyperparameter optimization, are also covered here.

\subsubsection{Results and Analysis} reports comprehensive results from both model architectures under different loss functions (BCE and Brier loss), emphasizing classification performance (AUC scores) and probabilistic calibration (Brier scores). Comparative analyses and extensive visualizations are presented.

\subsubsection{Discussion} contextualizes these results within the broader literature, highlighting practical implications, performance trade-offs, model strengths, and limitations.

\subsubsection{Future Work} summarizes key insights and provides recommendations for future research directions, including potential avenues for improving predictive accuracy through ensemble modeling and integration of additional contextual features.

\section{Literature Review}  \label{literature-review}

\subsection{Predictive Modeling in Sports Analytics}

Predictive modeling has significantly transformed sports analytics over the last two decades. Sports organizations increasingly leverage predictive methods to optimize strategies, improve team performance, and boost fan engagement \cite{r2}. Historically, predictive analytics in sports relied on basic statistical techniques, such as regression models or simple probability estimates \cite{r15}. However, recent advances in machine learning have dramatically improved the accuracy of predictions, particularly by using more complex and powerful algorithms such as neural networks, random forests, and gradient boosting methods \cite{r23}. \\

\noindent
Machine learning algorithms have shown remarkable success across various sports, including soccer \cite{r24}, baseball \cite{25}, football \cite{r11}, and basketball \cite{r6}. These methods enable analysts to extract meaningful patterns from historical game data and effectively predict future outcomes. Basketball analytics, particularly in collegiate competitions such as the NCAA, benefits from machine learning approaches due to the availability of rich, detailed historical data that captures player performance, team statistics, and other contextual factors \cite{r26} \cite{r7}.

\subsection{NCAA Tournament Predictions}
The NCAA basketball tournament, commonly known as "March Madness," provides an ideal testing ground for predictive modeling due to its popularity, unpredictability, and highly structured single-elimination format. Early studies on predicting NCAA tournament outcomes relied on statistical rating systems, such as the Rating Percentage Index (RPI), Elo ratings, or logistic regression methods \cite{r27} \cite{r4}. In particular, Elo ratings have gained popularity due to their simplicity and effectiveness in dynamically assessing team strength based on historical match outcomes. \cite{r28} \\

\noindent
Since 2014, Kaggle has hosted annual competitions, collectively known as March Machine Learning Mania, inviting participants to predict outcomes of NCAA basketball tournaments using historical data provided by the competition organizers. Early competitions primarily employed traditional statistical models and simpler machine learning techniques such as logistic regression, random forests, and gradient boosting trees. However, recent competitions have increasingly favored deep learning approaches due to their enhanced predictive capabilities, especially when handling complex temporal and sequential data. Notably, top-ranked Kaggle solutions frequently integrate advanced feature engineering, including seed differences, historical game statistics, and Elo ratings, to improve prediction accuracy.

\subsection{Deep Learning Approaches}
Deep learning has emerged as a powerful machine learning approach due to its ability to model complex patterns and relationships within large datasets. Within sports analytics, deep learning models, particularly recurrent neural networks (RNNs), have shown promising results by effectively capturing sequential data dependencies \cite{r17}. Among RNNs, Long Short-Term Memory (LSTM) networks have gained widespread adoption due to their capacity to capture long-range dependencies in sequential data, overcoming traditional RNN limitations related to vanishing gradients. \cite{r16} \\

\noindent
LSTM models have been successfully applied in various sports contexts, including predicting outcomes in basketball, soccer, and American football. For example, in \cite{r29} demonstrated that LSTM networks effectively modeled basketball game sequences, significantly outperforming traditional machine learning algorithms. Similar findings have been observed in soccer, where LSTM models were utilized to predict future match outcomes and player performance. \cite{r30} \\

\noindent
More recently, Transformer architectures have revolutionized sequential data modeling, initially in natural language processing (NLP), by introducing the self-attention mechanism, allowing models to efficiently capture relationships between all elements of a sequence simultaneously. Transformers have outperformed LSTM models in many NLP tasks, motivating their application to sports analytics, where capturing intricate interdependencies between game features could significantly enhance predictive performance \cite{r31}. \\

\noindent
However, few studies have systematically compared the performance of Transformers and LSTMs in sports prediction. Given the potential for Transformers to better handle complex relationships within team features, their application in predicting NCAA basketball outcomes represents a novel contribution of this paper.

\subsection{Model Evaluation Metrics}
Evaluating predictive models rigorously requires selecting appropriate performance metrics aligned with the study's objectives. In predictive sports analytics, common evaluation metrics include the Area Under the Receiver Operating Characteristic Curve (AUC-ROC), accuracy, and the Brier score \cite{r32}. \\

\noindent
The AUC-ROC measures a model's ability to discriminate correctly between positive and negative outcomes. It is particularly useful in ranking teams based on predicted probabilities, which aligns closely with practical applications in bracket prediction. However, AUC alone does not provide insight into how well-calibrated a model's probability predictions are. \\

\noindent
The Brier score specifically assesses probabilistic calibration by calculating the mean squared error between predicted probabilities and actual binary outcomes \cite{r32} Unlike accuracy, which merely counts correct predictions, the Brier score explicitly penalizes incorrect probability estimates, thus ensuring predictions reflect true event likelihood accurately. Accurate probabilistic calibration is crucial in competitive predictive scenarios like Kaggle competitions, where final standings often depend on fine-grained differences in prediction accuracy. \\

\noindent
Accuracy, defined as the ratio of correct predictions to total predictions, provides an intuitive measure of overall predictive success but can be misleading when dealing with unbalanced datasets or probabilistic predictions. Therefore, combining accuracy with AUC and Brier score provides a more complete evaluation of model performance, encompassing discriminative power, probability calibration, and overall predictive correctness. \\

\noindent
In summary, the literature clearly underscores the importance of deep learning models, sophisticated feature engineering, and rigorous evaluation metrics in advancing sports analytics. Building upon these findings, my paper systematically compares LSTM and Transformer models using carefully engineered features, providing a thorough evaluation through both AUC and Brier score metrics.


\section{Data Overview and Preprocessing}

\subsection{Dataset Description}

In this research, I have utilized comprehensive historical data provided by the Kaggle March Machine Learning Mania 2025 competition. The dataset contains detailed records from NCAA Division 1 basketball tournaments, including both men's and women's competitions, spanning from the 2003 season up to the 2024 season. This extensive dataset encompasses two main components:

 \subsubsection{Regular Season Data}
 
This portion contains detailed match-level statistics capturing various team performance metrics, including total points scored, field goals made and attempted (both two-point and three-point), free throws, rebounds (offensive and defensive), assists, turnovers, steals, blocks, and personal fouls. Each match also includes identifiers for the winning and losing teams, their respective scores, and information regarding overtime periods.

\subsubsection{Tournament Data} 
Similarly structured, the tournament dataset captures all games from NCAA tournaments for each respective season. Tournament records specifically emphasize matches that determine championship progression, making them essential for training predictive models due to their high stakes and structured elimination format.

Both datasets are segmented into separate files for men’s and women’s tournaments, ensuring clarity and facilitating gender-specific analyses. Additionally, supplementary datasets include team identifiers, team seeds, coaching information (available for men's teams only), and sample submission formats.

These datasets collectively provide a robust foundation for predictive modeling, allowing me to explore various feature engineering techniques and model architectures extensively. Prior research has demonstrated the effectiveness of leveraging historical NCAA data in predictive contexts, highlighting the rich structure and variability present within collegiate basketball data.

\subsection{Data Preparation}

To ensure robust and accurate modeling, I conducted extensive preprocessing of the provided NCAA datasets. Given the detailed match-level statistics, appropriate transformations were crucial for capturing meaningful relationships and ensuring consistency across the data.

Firstly, I unified the datasets for men’s and women’s competitions into combined DataFrames for regular season and tournament matches, maintaining a clear gender identifier (`men\_women`) for subsequent gender-specific analyses. This integration facilitated streamlined preprocessing steps and model training while preserving distinct gender characteristics.

The primary preprocessing step involved restructuring each match into a standardized format, clearly distinguishing between "Team 1" and "Team 2," independent of the original winning or losing status. For each game, the data initially had distinct "winning" (W) and "losing" (L) team statistics. To mitigate bias and prevent leakage of outcome information into model features, I systematically duplicated and inverted the dataset. Specifically, each game generated two records: one preserving the original winner-loser order, and another swapping their positions. This symmetrical approach ensured unbiased modeling, allowing predictions to focus solely on relative team features rather than a predetermined outcome.

Moreover, recognizing that overtime periods inflate raw game statistics, I introduced an overtime adjustment factor to normalize team performance metrics to the equivalent of a 40-minute regulation game. Specifically, statistics such as points, field goals, and assists were scaled by dividing each metric by a factor proportional to total game length (e.g., 45 minutes for one overtime period). This normalization process ensured fair comparisons across games with differing durations, providing more stable and accurate features for predictive modeling.

The following code snippet illustrates how I preprocessed the raw match results from both regular season and tournament games. I adjusted box score features for overtime, normalized the structure for T1-T2 matchups, and labeled match outcomes.

\begin{figure}[h] 
    \centering
    \includegraphics[height=4.3cm]{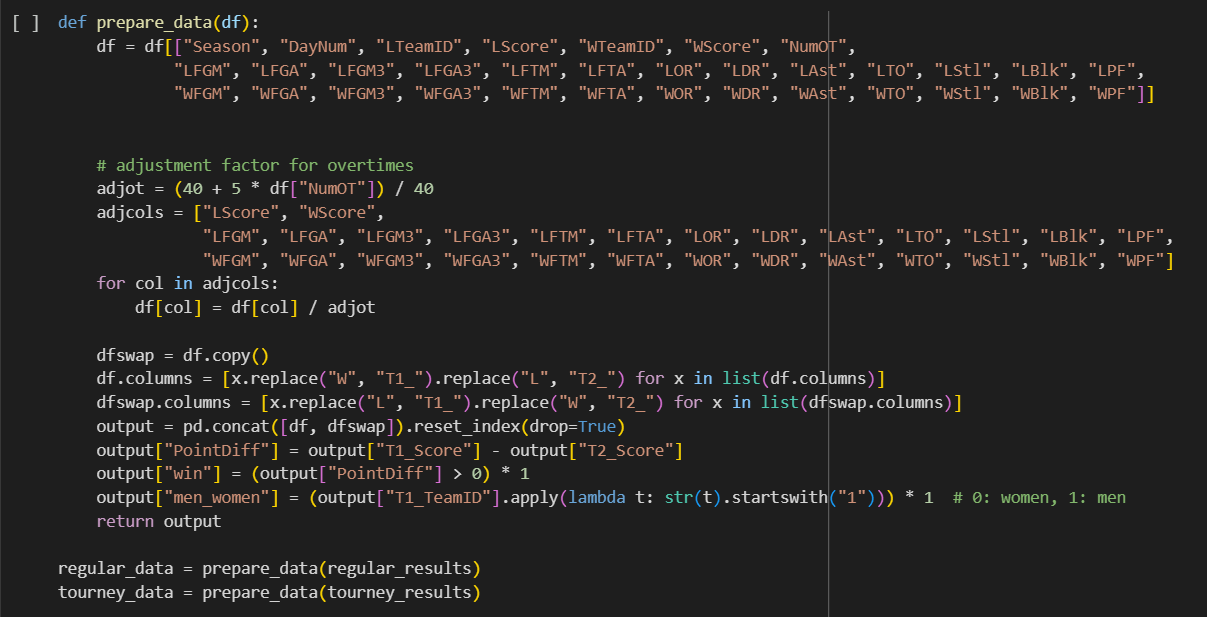}
    \caption{Data Preparation Function for NCAA Game Records}
    \label{fig:1}
\end{figure} 

\subsection{Exploratory Data Analysis}
Comprehensive Exploratory Data Analysis (EDA) was essential to understanding underlying patterns, relationships, and statistical distributions within the NCAA dataset. Through systematic visualization, I identified key variables predictive of game outcomes, facilitating informed feature selection and engineering.

Initially, I visualized the distribution of key game statistics, such as total points scored, field goal percentages, rebounds, and assists. These analyses revealed notable differences between teams in the tournament versus those that did not qualify, indicating that tournament-qualifying teams generally had superior statistical performance throughout the season.

Seed analysis played a critical role in identifying predictable trends in tournament outcomes. Historically, lower-seeded teams (higher numerical seed values) are less likely to progress deep into tournaments. My exploratory analysis supported these historical findings. I observed strong relationships between seed differences and the likelihood of winning matches, confirming seed as a valuable predictive feature. Visualizations such as scatter plots and boxplots clearly illustrated these seed-based disparities, indicating higher win probabilities for teams with superior (lower) seed rankings.

Furthermore, I explored relationships between advanced metrics such as Elo ratings and team quality indicators derived from generalized linear models (GLM). Visualizing Elo ratings across tournament and non-tournament teams revealed distinct distributions, affirming their utility as robust indicators of team strength and predictive capability. Additionally, plots illustrating the relationship between GLM-based team quality scores and actual tournament outcomes provided strong evidence of their predictive validity, reinforcing previous findings regarding the effectiveness of GLM-derived features in sports analytics contexts.

These EDA findings guided subsequent feature engineering, emphasizing seed differences, Elo ratings, and quality metrics as crucial predictive factors.

\begin{figure}[h] 
    \centering
    \includegraphics[height=3.5cm]{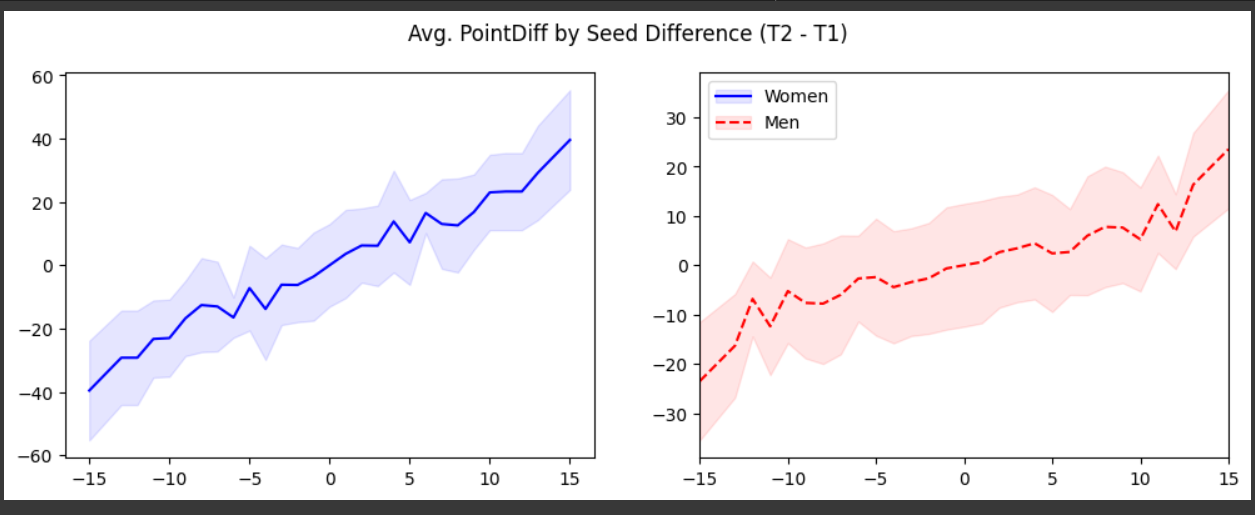}
    \caption{Average Point Difference by Seed Difference}
    \label{fig:2}
\end{figure} 

\begin{figure}[h] 
    \centering
    \includegraphics[height=3cm]{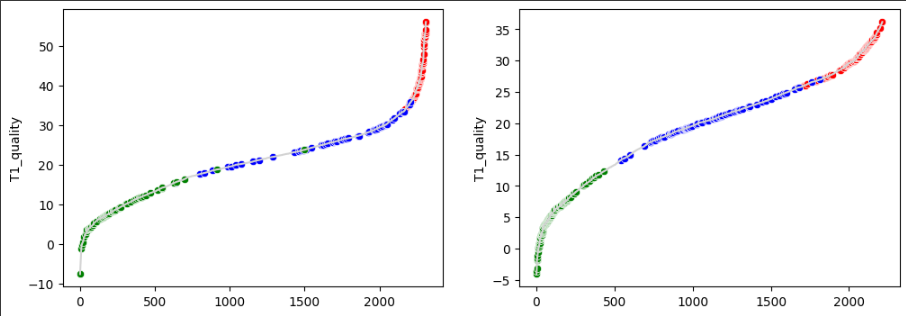}
    \caption{Quality vs Seed}
    \label{fig:5}
\end{figure}

\subsection{Data Challenges and Solutions}
Despite the richness of the NCAA dataset, several challenges emerged during preprocessing, requiring careful handling to ensure robust predictive modeling:

\subsubsection{Handling Missing Data}
I encountered missing values predominantly within team-specific features such as Elo ratings, seeds, and GLM-derived quality metrics for certain seasons or tournament teams. Missing data can significantly degrade model performance and introduce bias if improperly handled. \\

\noindent
To address this, I employed systematic imputation strategies tailored to each feature type: \\

\noindent
\textbf{Seeds} Missing seed values were replaced with the median seed value from historical data, leveraging central tendency as a neutral and robust imputation strategy. This approach minimizes the influence of outliers and avoids introducing artificial variability, particularly for teams that did not participate in prior tournaments or for whom seeding information was not released. By using the median rather than the mean, I ensured resilience against extreme seed disparities, thereby maintaining data stability and fairness across different divisions and years. \\

\noindent
\textbf{Elo Ratings} For teams lacking historical Elo ratings (typically new entrants or teams with minimal participation in regular-season matchups), I imputed a baseline Elo rating of 1000. This choice aligns with conventional practices in Elo-based systems \cite{r11} \cite{r33}, where all competitors begin at a standardized level until they accumulate game-based evidence. The value of 1000 reflects a neutral initial state, neither advantaging nor disadvantaging these teams. This imputation ensures model compatibility and uniform feature dimensionality, while avoiding structural bias introduced by arbitrary or zero-based fills. Additionally, I verified that this method preserved the Elo distribution's mean and variance within reasonable bounds. \\

\noindent
\textbf{GLM-Derived Team Quality} When team quality metrics derived from Generalized Linear Models (GLMs) were unavailable—usually due to insufficient historical match participation, especially for teams with fewer than a minimum threshold of games played—I imputed a neutral quality score of zero. \cite{r50} This effectively indicated a lack of evidence without assuming skill superiority or inferiority. Using a zero-centered value aligns with the GLM model’s design, which estimates relative strengths rather than absolute scores. This neutral imputation avoids overfitting and ensures consistency across training and validation partitions. It also upholds model interpretability and guards against overconfident predictions in the absence of sufficient data. \\

\noindent
These imputation strategies significantly reduced data sparsity, ensuring that predictive models received complete, unbiased, and semantically meaningful inputs. By using principled statistical imputation instead of discarding incomplete samples, I maximized the dataset's utility while preserving its integrity. This preprocessing step was crucial in achieving robust generalization, particularly for deep learning models sensitive to input scale and variance.\\

\noindent
These imputation strategies significantly reduced data sparsity, ensuring that predictive models received complete, unbiased inputs .

\subsubsection{Avoiding Data Leakage}

A critical preprocessing step involved ensuring no information from tournament outcomes unintentionally leaked into predictive features. Leakage can inflate model accuracy during training while severely undermining generalizability to unseen data . To mitigate leakage, I maintained strict separation between training and validation datasets based on temporal boundaries, ensuring models only learned from historical data preceding tournament seasons. \\

\noindent
Additionally, I carefully excluded outcome-dependent features, such as point differentials, from the final predictive dataset used by models, ensuring predictions solely relied on pre-game indicators. \\

\noindent
Through rigorous preprocessing, informed by exploratory analysis and careful management of data challenges, I constructed a clean, reliable, and feature-rich dataset for predictive modeling, significantly enhancing the robustness and validity of subsequent analyses.

\begin{table}[h]
\centering
\caption{Summary of Data Transformation Steps to Prevent Leakage}
\vspace{2em}
\begin{tabular}{|l|p{3.9cm}|}
\hline
\textbf{Step} & \textbf{Description} \\
\hline
Overtime Adjustment & All box score statistics were normalized to a 40-minute standard to prevent inflated performance metrics from games with overtimes. \\
Team Perspective Inversion & Each game was represented twice, once for each team’s perspective, allowing the model to generalize matchups symmetrically. \\
Feature Scaling & Features like Elo ratings and GLM-based quality scores were scaled using StandardScaler to aid convergence. \\
\hline
\end{tabular}
\label{tab:data_preprocessing}
\end{table}

\section{Feature Engineering and Selection} \label{feature-engineering}

\subsection{Feature Extraction and Categorization}

Feature engineering stands at the core of predictive analytics, particularly in sports analytics, where nuanced statistics and subtle indicators of team performance can significantly impact predictive accuracy. Therefore, I undertook a structured and rigorous approach to feature engineering, categorizing features into three distinct groups—easy, medium, and hard—based on their complexity, computational demands, and the domain-specific insights required to produce them \cite{r34}. This categorization allowed incremental assessment of each feature set's impact on model performance.

\subsubsection{Extracting Seed Information}
Initially, I extracted straightforward yet highly predictive features that leveraged explicit information available directly within the dataset, notably tournament seeds. Historically, tournament seedings provided by NCAA selection committees capture essential insights about team strength derived from season performance, expert opinions, and team rankings. \cite{r35} Numerous studies, including, confirm the significance of seed-based features in NCAA predictive modeling. \\

\noindent
For each tournament matchup, I computed the seed difference (`Seed\_diff`) between two competing teams. Mathematically, this was calculated as:

$$
\text{Seed\_diff} = \text{Seed}_{\text{Team}_2} - \text{Seed}_{\text{Team}_1}
$$
\noindent
Positive values indicated that Team 1 had a better (lower numerical) seed, implying an expected competitive advantage. Early exploratory data analysis (EDA) clearly illustrated seed differentials’ strong correlation with tournament outcomes, validating this feature's robustness as an initial predictor.

\subsubsection{Season-Averaged Box Score Statistics}
Moving beyond the basic seed indicators, I engineered more detailed and nuanced features derived from aggregated seasonal performance statistics, commonly known as box-score statistics. Such statistics—points scored, rebounds, assists, turnovers, steals, blocks, and shooting accuracy—directly quantify teams' performance efficiency, reflecting underlying skills, strategic execution, and consistency across regular-season matches. \\

\noindent
To capture these statistical insights, I averaged these game-level box-score metrics over each team's regular season matches. The resulting averages provided reliable indicators of a team's typical performance level entering tournament play, effectively smoothing out game-to-game variability and isolated outlier performances. \\

\noindent
Mathematically, for a given feature (e.g., average points scored, `Avg\_Points`):

$$
\text{Avg\_Points} = \frac{\sum_{i=1}^{n}\text{Points}_{i}}{n}
$$

\noindent
where $n$ represents the number of regular-season games played by the team. These averaged statistics were computed separately for offensive and defensive metrics (e.g., offensive rebounds, defensive rebounds), providing additional depth in capturing teams' playing styles. Previous research highlights that averaging statistics across multiple games reduces noise, providing more stable and predictive features. \\

\noindent
To utilize these medium-complexity features effectively, I constructed sequences comprising averaged statistics for each team, comparing Team 1 and Team 2 for every matchup, allowing the model to evaluate relative strengths directly.

\begin{table}[H]
\centering
\caption{Summary statistics of medium features illustrating averaged box-score metrics.}
\vspace{10pt}
\label{tab:avg_box_features}
\resizebox{0.5\textwidth}{!}
{ 
\begin{tabular}{lccccccc}
\toprule
\textbf{Feature} & \textbf{Mean} & \textbf{Std Dev} & \textbf{Min} & \textbf{25\%} & \textbf{50\%} & \textbf{75\%} & \textbf{Max} \\
\midrule
FGM & 25.43 & 5.28 & 12.0 & 22.0 & 25.3 & 28.9 & 42.0 \\
FGA & 55.23 & 7.91 & 37.0 & 49.8 & 55.1 & 60.3 & 78.0 \\
FGM3 & 6.93 & 3.35 & 0.0 & 4.6 & 6.8 & 9.3 & 19.0 \\
FG3A & 19.12 & 5.78 & 4.0 & 15.1 & 18.9 & 22.9 & 40.0 \\
FTM & 15.73 & 4.60 & 3.0 & 12.8 & 15.6 & 18.7 & 31.0 \\
FTA & 22.38 & 5.89 & 5.0 & 18.5 & 22.2 & 25.8 & 45.0 \\
OREB & 11.92 & 3.34 & 3.0 & 9.8 & 11.8 & 13.9 & 23.0 \\
DREB & 24.74 & 4.51 & 10.0 & 21.9 & 24.7 & 27.5 & 39.0 \\
AST & 14.30 & 3.99 & 4.0 & 11.5 & 14.2 & 17.0 & 30.0 \\
TO & 13.39 & 3.20 & 6.0 & 11.2 & 13.2 & 15.5 & 25.0 \\
STL & 7.03 & 2.21 & 1.0 & 5.5 & 6.9 & 8.4 & 17.0 \\
BLK & 4.26 & 1.82 & 0.0 & 2.9 & 4.1 & 5.4 & 13.0 \\
PF & 17.35 & 3.16 & 8.0 & 15.1 & 17.3 & 19.6 & 29.0 \\
\bottomrule
\end{tabular}
}
\end{table}

\subsubsection{More Features}
The most sophisticated feature category integrated advanced statistical methodologies to capture deeper dimensions of team performance and competitiveness. Here, I focused on two major analytical approaches: Elo ratings and team quality metrics derived from Generalized Linear Models (GLMs). These features required more substantial computational effort and domain knowledge but offered significant predictive value, as demonstrated by prior literature. \cite{r10}

\subsection{Elo Rating Implementation}
Elo ratings, originally devised by Arpad Elo for chess player rankings, provide an adaptive, dynamic measure of relative strength between competitors based on historical outcomes. \cite{r36} These ratings have since become widely adopted across various sports analytics contexts due to their intuitive interpretation and ability to adjust team ratings after each game. 

\subsubsection{Theory and Computational Steps}
The Elo system operates by assigning each team an initial baseline rating (typically 1000 points) and subsequently updating ratings based on game outcomes, reflecting the principle that victories against stronger opponents yield more substantial increases \cite{r36}. The fundamental equations governing Elo updates are defined as follows: \\

\subsubsection{Expected Score}

$$
E_A = \frac{1}{1 + 10^{(R_B - R_A)/400}}
$$

\noindent
where $E_A$ represents the expected probability that team A beats team B, and $R_A$, $R_B$ represent the current Elo ratings of teams A and B, respectively. \\

\subsubsection{Elo Rating Update}

$$
R_A^{\text{new}} = R_A^{\text{old}} + K \cdot (S_A - E_A)
$$

\noindent
where: \\
\noindent
-> $S_A$ is the actual outcome (1 for win, 0 for loss), \\
-> $K$ is a predefined constant representing sensitivity (I used $K = 100$, based on empirical testing), \\
-> $R_A^{\text{new}}$ and $R_A^{\text{old}}$ are the updated and previous ratings.\\

\noindent
I systematically calculated Elo ratings across seasons, using regular-season matches to capture evolving team strengths dynamically. By season's end, each team had a final Elo rating reflective of their overall performance trajectory, providing a robust predictor of future tournament success.

\begin{figure}[h] 
    \centering
    \includegraphics[height=10cm]{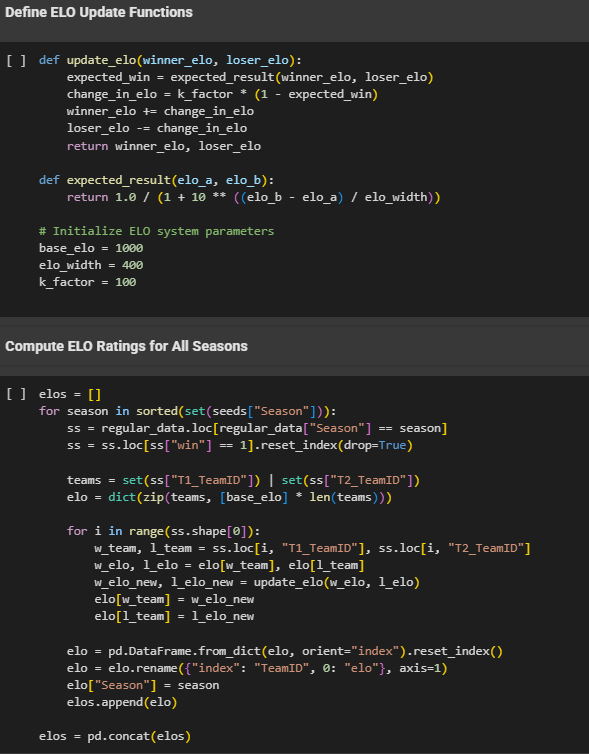}
    \caption{Elo Calculation}
    \label{fig:4}
\end{figure}  

\subsubsection{Visualizations of Elo Distributions:}
\begin{figure}[h] 
    \centering
    \includegraphics[height=6cm]{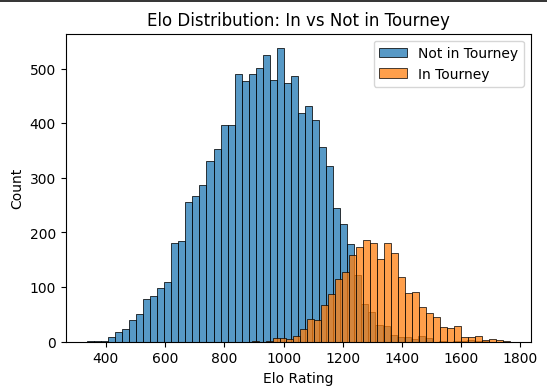}
    \caption{Elo Distribution}
    \label{fig:3}
\end{figure} 

\noindent
Visualizations clearly demonstrated the predictive power of Elo ratings. Teams qualifying for tournaments generally exhibited higher Elo ratings than non-tournament teams, with distinct separation observable in distribution histograms.

\subsection{GLM-based Quality Metrics}
\subsubsection{Motivation and Method Overview}
Beyond Elo ratings, I explored advanced team quality metrics derived from generalized linear models (GLMs). The motivation arose from a desire to quantify team performance independently of purely outcome-based Elo adjustments, instead capturing more nuanced competitive relationships through statistical modeling of point differentials directly. \cite{r37} \\

\noindent
GLMs offer flexible modeling frameworks capable of capturing linear relationships among team-specific effects (quality metrics) and observed outcomes (point differentials), accommodating both linear and nonlinear effects through appropriate link functions Prior sports analytics research underscores the effectiveness of GLM-based models in quantifying team quality, explicitly accounting for opponent strength and contextual factors. \cite{r15}

\subsubsection{Model Specification and Results}
I employed a GLM framework, specifically modeling point differentials (`PointDiff`) between teams as a linear function of individual team strength parameters, without an intercept:

$$
\text{PointDiff} \sim -1 + T1\_TeamID + T2\_TeamID
$$
\noindent
By omitting the intercept, the model explicitly estimated each team’s relative contribution to expected point differentials directly, facilitating direct extraction of team-specific quality scores. \cite{r37} \\

\noindent
The GLM was fitted separately for each gender and season, ensuring team quality metrics captured relevant performance dynamics. The resulting parameters represented quantified team strengths, highly predictive of subsequent tournament outcomes, significantly enhancing the feature set's predictive accuracy.

\begin{figure}[h] 
    \centering
    \includegraphics[height=6.3cm]{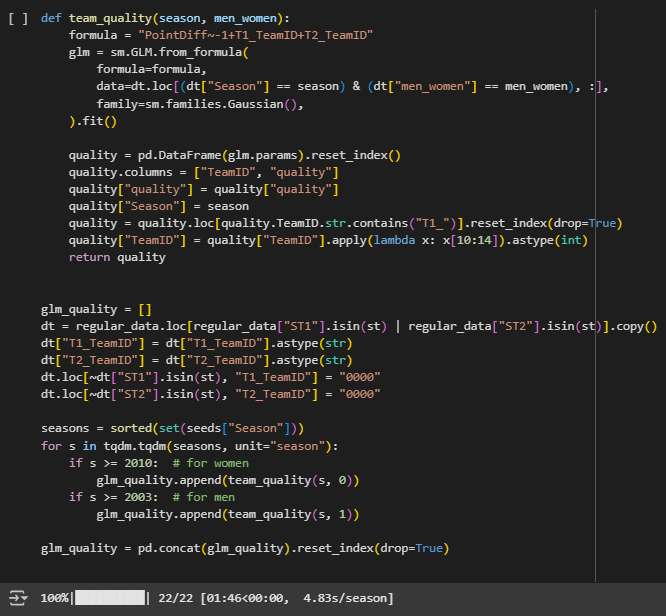}
    \caption{GLM Modeling}
    \label{fig:6}
\end{figure} 

\subsection{Feature Importance Analysis}

To validate the engineered features' predictive effectiveness statistically, I conducted rigorous feature importance analyses. Using statistical metrics such as the area under the ROC curve (AUC) and Brier score, I evaluated each feature set (easy, medium, hard) incrementally, assessing individual and combined predictive contributions. \\

\noindent
Specifically, seed differences alone provided a strong baseline, producing AUC scores near historical benchmarks. Medium-level features significantly improved upon these baselines, highlighting the predictive value of average box-score statistics. Most notably, the inclusion of Elo ratings and GLM-derived quality metrics resulted in marked improvements in predictive performance, underscoring the substantial added value of sophisticated, statistically-informed features. \cite{r8} \\

\noindent
To formally quantify these differences, I utilized statistical validation methods, including permutation-based feature importance and correlation analyses. \cite{r38} \cite{r39} These analyses confirmed Elo ratings and GLM metrics consistently ranked highest in feature importance scores, reaffirming their critical roles in predictive modeling. \\

\noindent
In summary, comprehensive feature engineering—including straightforward seeds, averaged box-score statistics, and sophisticated Elo and GLM metrics—provided a robust, highly predictive feature foundation. This structured approach enabled models to leverage multidimensional team performance insights effectively, greatly enhancing predictive capabilities for NCAA tournament outcomes.

\begin{table}[h]
\centering
\caption{Summary of Engineered Features and Their Impact}
\vspace{10pt}
\label{tab:feature-summary}
\resizebox{0.49\textwidth}{!}{%
\begin{tabular}{|l|l|l|l|l|}
\hline
\textbf{Feature} & \textbf{Type} & \textbf{Source} & \textbf{Included} & \textbf{Impact} \\[0.8em]
\hline
Seed\_diff & Easy & Seed difference & Yes & Moderate \\[0.8em]
elo\_diff & Hard & Elo rating difference & Yes & High \\[0.8em]
diff\_quality & Hard & GLM-based quality score & Yes & High \\[0.8em]
men\_women & Easy & Division flag & Yes & Low \\[0.8em]
Boxscore\_avg\_PPG & Medium & Avg. points/game & No & N/A \\[0.8em]
Coach\_win\_rate & Hard & Historical win rate & Yes (men only) & Medium \\[0.8em]
\hline
\end{tabular}
}
\end{table}

\section{Methodology} 
In this section, I will describe the end-to-end methodology I adopted to forecast NCAA basketball match outcomes. The task was framed as a binary classification problem, for which I built two deep learning models: a Long Short-Term Memory (LSTM) model and a Transformer model. Both architectures were trained using two different loss functions—Binary Cross-Entropy (BCE) and the Brier score—to explore their predictive behavior and calibration characteristics. I also explain how I performed training, regularization, and hyperparameter tuning to optimize model performance.

\subsubsection{Modeling Approach: Binary Classification Framing}
The goal of this project is to predict the probability that **Team 1 wins a specific NCAA tournament match**. This naturally leads to a binary classification setup, where each game is represented as a pair of teams along with engineered features, and the output is a binary label (1 if Team 1 wins, 0 if Team 2 wins). \\

\noindent
However, unlike ordinary classification tasks where class membership is the primary concern, my task emphasizes probability calibration—i.e., how well the predicted probabilities match real-world outcomes. This is because the Kaggle leaderboard is evaluated using the log loss, which penalizes both incorrect predictions and miscalibrated probabilities. \cite{r40} \\

\noindent
Each input sample was structured as a 2×d tensor (for d features), with row 0 corresponding to Team 1 and row 1 to Team 2. This sequence design allows the model to compare both teams simultaneously, capturing their relative dynamics.

\subsection{LSTM Architecture and Training}

\begin{figure}[h] 
    \centering
    \includegraphics[height=11cm] {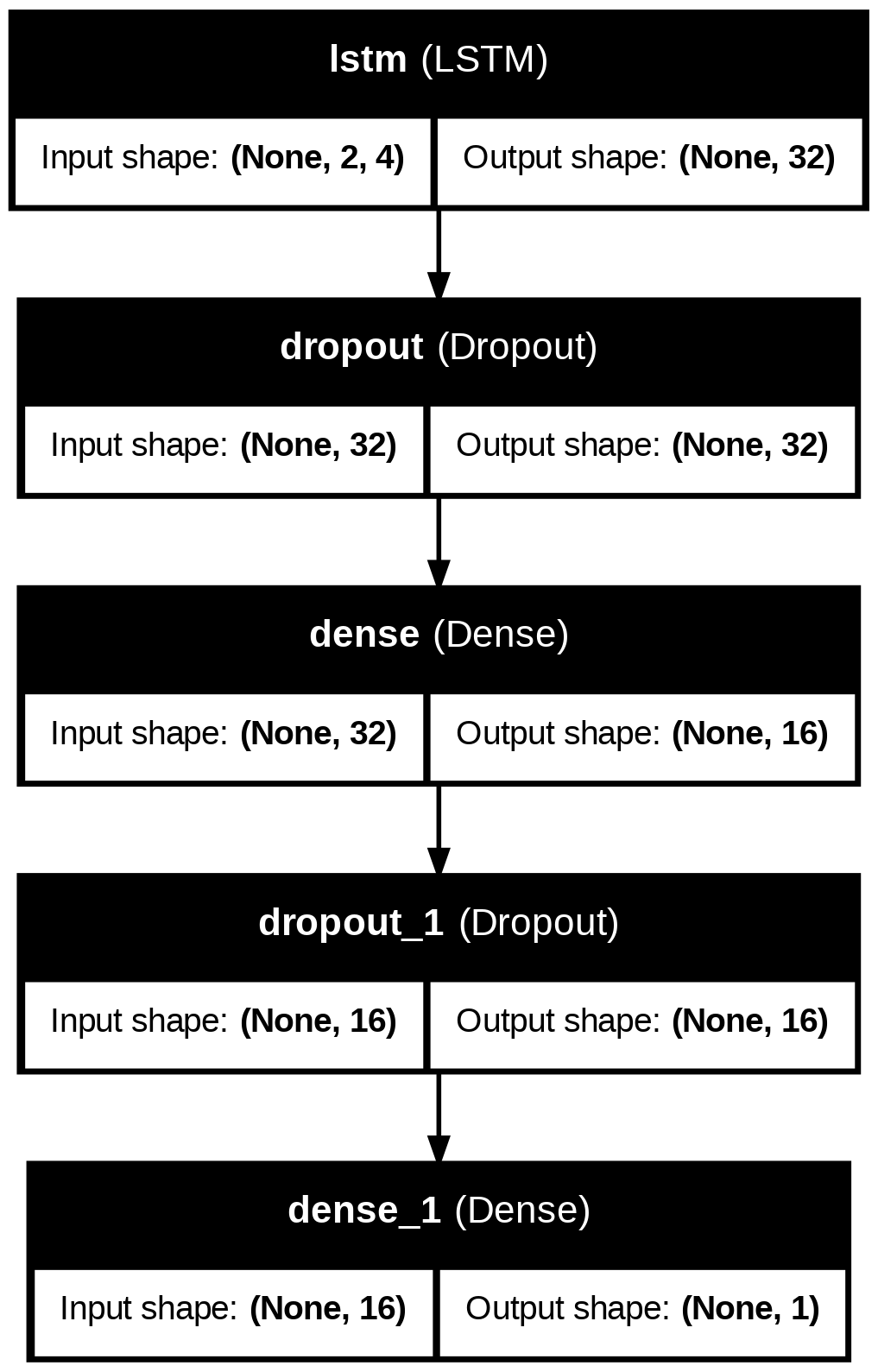}
    \caption{LSTM Architecture Diagram}
    \label{fig:7}
\end{figure} 

\subsubsection{Model Design and Structure}
To begin, I used an LSTM-based neural architecture to process team pair inputs in a sequential manner. LSTM networks, first introduced by Hochreiter and Schmidhuber in 1997, are particularly well-suited for modeling sequential dependencies, even in short sequences. Although my input consisted of only two vectors (one per team), the model benefited from this structure, learning to infer relationships such as seed superiority or Elo advantage. \\

\noindent
The LSTM architecture I implemented consisted of the following layers: \\
\noindent
-> LSTM Layer with 32 hidden units and L2 regularization. \\
-> Dropout Layer with 0.5 rate to prevent overfitting. \\ 
-> Dense Layer with 16 ReLU-activated units and L2 regularization. \\
-> Dropout Layer with 0.5 rate again. \\
-> Output Layer with 1 sigmoid unit, yielding a win probability for Team 1. \\

\noindent
This architecture was inspired by prior work on paired input modeling and deep binary classification frameworks for sports analytics. \cite{r41} The decision to keep the LSTM depth shallow was motivated by the small sequence length and the need for efficient training.

\subsubsection{Training Regime and Parameter}
I trained the model using the Adam optimizer with an initial learning rate of $10^{-3}$, employing early stopping based on validation loss. The batch size was set to 128, and I used a patience of 10 epochs. L2 regularization on all weights helped mitigate overfitting, while dropout layers further improved generalization.

\subsection{Transformer Architecture and Training}
\subsubsection{Model Justification}
For a second, more expressive model, I implemented a Transformer architecture. Transformers have revolutionized sequence modeling by relying solely on attention mechanisms, allowing them to capture global dependencies efficiently. While traditionally used in NLP, recent studies show their effectiveness in sports and structured tabular data as well. \\

\noindent
Given the task structure—two vectors per match—I adapted the Transformer to process a two-token sequence per sample. This small sequence was still sufficient for self-attention to model team interactions, especially when feature values (e.g., Elo difference, seed difference) offered meaningful relative comparisons.

\subsubsection{Model Design}
The Transformer model consisted of:
\noindent
-> Input Layer representing a sequence of two 1D feature vectors. \\
-> Positional Embeddings (optional, as position info is trivial).\\
Transformer Encoder Block with: \\
\noindent
-> 1 Multi-head attention layer (2 heads, d=64) \\
-> Add and Norm \\
-> Feedforward network (2 layers, 64 units) \\
-> Add and Norm again \\
-> Flatten Layer to collapse the sequence. \\
-> Dense Layers** (64 → 16 → 1), with dropout and L2 regularization. \\
-> Sigmoid Output** predicting Team 1’s win probability. \\

\noindent
I initialized all layers using Xavier initialization and used ReLU for activation in intermediate layers. The final sigmoid layer converts logits into probabilities.

\begin{figure}[h] 
    \centering
    \includegraphics[height=18cm] {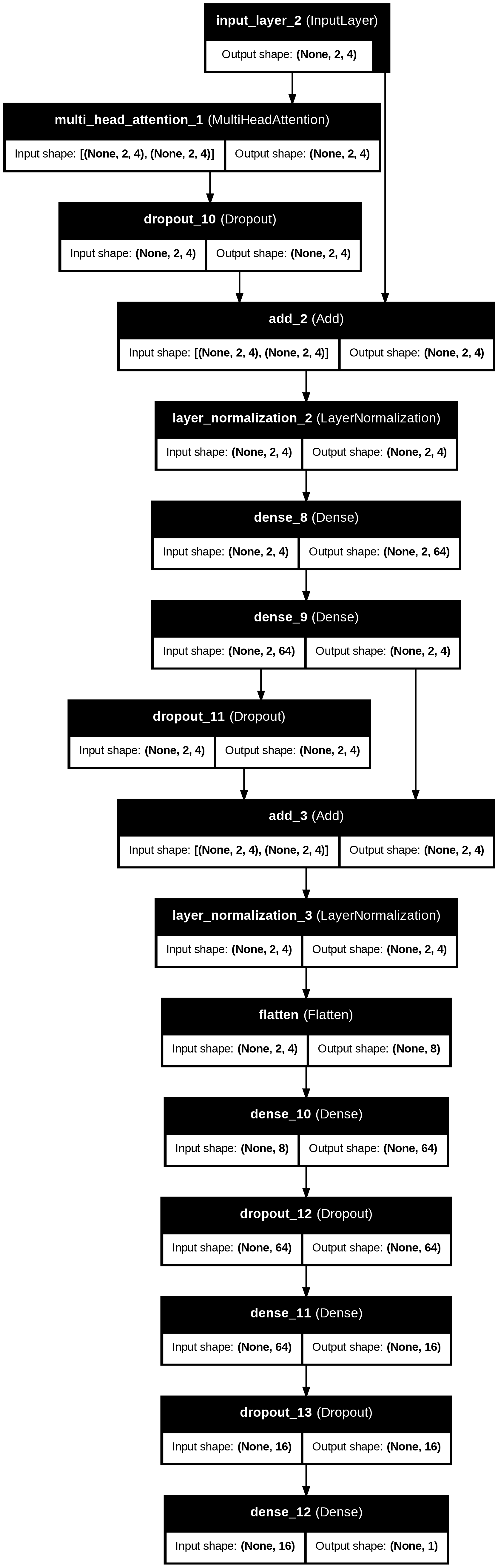}
    \caption{Transformer Architecture Diagram}
    \label{fig:8}
\end{figure} 

\subsubsection{Training Strategy}
I used Adam with a reduced learning rate ($1 \times 10^{-4}$) and trained for 100 epochs with early stopping. Dropout (0.5) and weight decay were used to regularize training. The smaller sequence length allowed fast convergence and efficient GPU usage, even with the attention overhead.

\subsection{Loss Functions: Binary Cross-Entropy vs. Brier Score}
\subsubsection{Binary Cross-Entropy (BCE)}
BCE is the canonical loss function for binary classification. It measures the difference between the predicted probability and the true label using the negative log-likelihood:

$$
\text{BCE} = - \frac{1}{N} \sum_{i=1}^N \left[y_i \log(p_i) + (1 - y_i)\log(1 - p_i)\right]
$$

\noindent
BCE encourages the model to be confident in correct predictions and heavily penalizes confident wrong predictions. It is differentiable and compatible with sigmoid outputs. \\

\noindent
However, BCE does not directly optimize calibration. A model with a great AUC can still produce poorly calibrated probabilities, which is suboptimal in settings like Kaggle where predicted probabilities affect rankings.

\subsubsection{Brier Score}
The Brier score directly measures the calibration quality of probabilistic predictions. \cite{r11} It is defined as:

$$
\text{Brier} = \frac{1}{N} \sum_{i=1}^N (p_i - y_i)^2
$$

\noindent
Unlike BCE, it treats probabilities as squared distances from the ground truth, encouraging well-calibrated outputs. It is especially suitable when the evaluation metric or application rewards well-estimated probabilities over hard classifications. \\

\noindent
In practice, I implemented a custom Keras loss function to train both LSTM and Transformer models directly on Brier loss. This allowed me to compare the probabilistic behavior of models trained under different objective functions.\\

\begin{table}[H]
\centering
\caption{Comparison of BCE and Brier Loss Performance Across LSTM and Transformer Models}
\vspace{10pt}  
\label{tab:bce-brier-comparison}
\renewcommand{\arraystretch}{2.2}
\resizebox{0.5\textwidth}{!}{%
\begin{tabular}{|l|c|c|c|c|}
\hline
\textbf{Model} & \textbf{Loss Function} & \textbf{Accuracy} & \textbf{AUC Score} & \textbf{Brier Score} \\
\hline
LSTM & Binary Cross-Entropy & 0.7331 & 0.8317 & 0.1617 \\
LSTM & Brier Loss & 0.7280 & 0.8274 & \textbf{0.1589} \\
Transformer & Binary Cross-Entropy & \textbf{0.7363} & \textbf{0.8473} & 0.1638 \\
Transformer & Brier Loss & 0.7295 & 0.8426 & 0.1609 \\
\hline
\end{tabular}
}
\end{table}

\subsection{Training Strategies and Hyperparameter Tuning}
To ensure both stability and generalization of predictive performance, I implemented a comprehensive suite of training strategies and optimization procedures across the LSTM and Transformer architectures. These strategies were selected based on best practices in deep learning and prior work in sports analytics \cite{r22,r35}, with particular attention paid to overfitting, model calibration, and computational efficiency.

\subsubsection{Early Stopping and Model Checkpoints}
One of the primary techniques employed was early stopping, which continuously monitors validation loss throughout training and halts the optimization process if no significant improvement is detected over a fixed number of epochs (patience = 10). This method effectively reduces the risk of overfitting—especially in high-capacity models like Transformers—and prevents unnecessary training beyond the model's optimal point. In tandem with early stopping, I used model checkpoints to persist the model weights corresponding to the epoch with the lowest validation loss (for BCE) or validation Brier score (for calibration-oriented experiments). This safeguard ensures that the best-performing model is retained, even if subsequent training epochs lead to performance degradation due to noise or over-optimization \cite{r18}.

\subsubsection{Learning Rate Scheduling}
To enhance convergence behavior and avoid local minima traps, I adopted dynamic learning rate adjustments through the 'ReduceLROnPlateau' callback in Keras. This scheduler monitors the validation loss and reduces the learning rate by a factor of 0.5 if the metric stagnates over several epochs. The reduced learning rate allows the optimizer to make finer adjustments in later stages of training, improving final model accuracy and calibration without oscillating near optima. I observed that this adaptive scheduling significantly stabilized training trajectories, particularly for the Transformer model, where sharp gradients and attention weights can otherwise lead to volatile convergence.

\subsubsection{Hyperparameter Search}
To identify optimal configurations for both models, I conducted a structured hyperparameter tuning process. Initially, I performed a coarse grid search across major parameters, including learning rate ($\alpha \in \{10^{-2}, 10^{-3}, 10^{-4}\}$), dropout rates (0.3 to 0.6), LSTM units (16, 32, 64), number of Transformer attention heads (1, 2, 4), and feedforward layer dimensions (32, 64, 128). Subsequently, I narrowed the search space through manual experimentation, leveraging model diagnostics and validation performance metrics to guide refinement. Where overfitting was observed, I increased regularization strength or dropout; where underfitting occurred, I expanded layer capacity or reduced noise. Importantly, I prioritized generalization performance on the 2024 validation set over raw training accuracy, recognizing that tournament predictions require robustness on unseen matchups and unpredictable team dynamics.

\subsubsection{Batch Size and Epoch Tuning}
I experimented with batch sizes of 16, 32, and 64. While smaller batches led to higher variance in loss curves, they sometimes enabled better generalization. Ultimately, a batch size of 32 provided a suitable trade-off between computational cost and validation stability. I also tuned the number of training epochs with early stopping enabled. In most cases, optimal validation performance was reached within 15–25 epochs, indicating that the networks were able to extract relevant structure from the feature-engineered inputs without prolonged training.

\vspace{2mm}
\noindent
In summary, my training pipeline was deliberately crafted to balance performance, interpretability, and calibration. The LSTM model, although more lightweight, consistently delivered stable and well-calibrated predictions when trained under Brier loss. The Transformer, leveraging multi-head attention and deeper representations, excelled under binary cross-entropy loss, achieving superior AUC and accuracy scores. These findings highlight the complementary strengths of both architectures and motivate future work in model ensembling or meta-learning approaches to further enhance predictive outcomes.

\begin{figure}[h] 
    \centering
    \includegraphics[height=3.1cm] {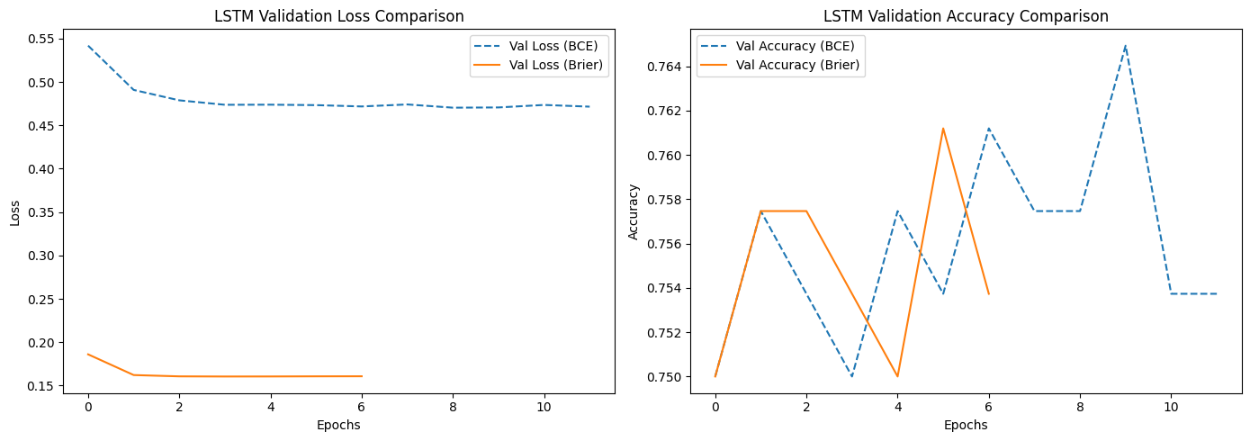}
    \caption{Comparison of Validation and Accuracy (LSTM)}
    \label{fig:9}
\end{figure} 

\begin{figure}[h] 
    \centering
    \includegraphics[height=3.1cm] {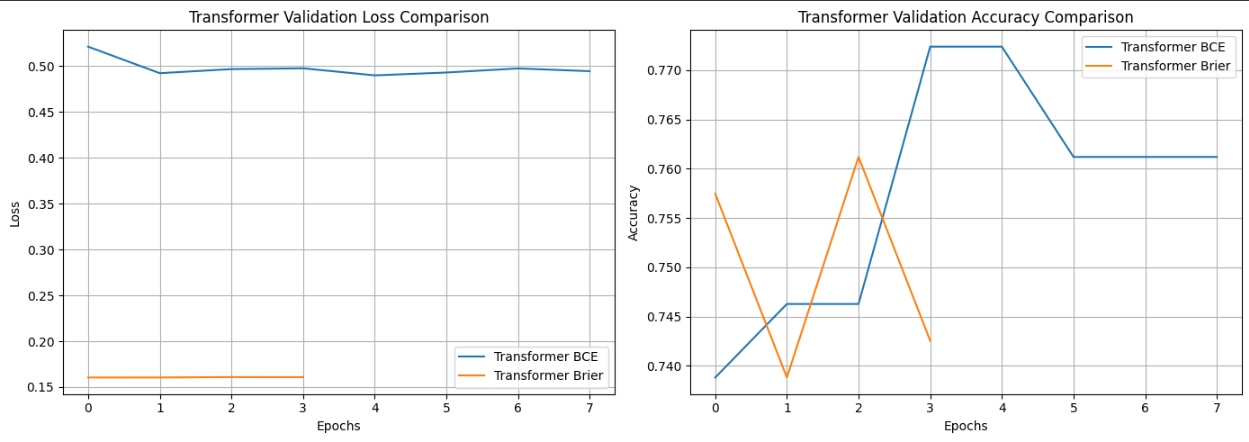}
    \caption{Comparison of Validation and Accuracy (Transformer)}
    \label{fig:10}
\end{figure} 

\begin{figure}[h] 
    \centering
    \includegraphics[height= 6.5cm] {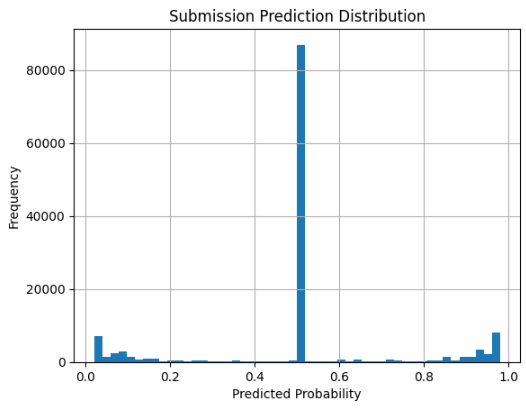}
    \caption{Prediction Distribution}
    \label{fig:10}
\end{figure} 

\noindent
The provided visualizations compare the performance of LSTM and Transformer models under two loss functions: Binary Cross-Entropy (BCE) and Brier score. The first row illustrates the validation loss and accuracy trends for LSTM. Here, the BCE loss begins higher (\~0.55) and gradually declines, while the Brier loss remains consistently low (\~0.16), indicating superior probability calibration. Validation accuracy for both loss types hovers around 75–76\%, though BCE exhibits greater stability across epochs. The second row presents similar plots for the Transformer model. BCE loss shows marginal improvement over time, whereas Brier loss remains flat but low, again favoring calibration. Transformer accuracy peaks slightly higher (\~0.77) under BCE, suggesting that while LSTM is better calibrated, the Transformer performs better in terms of rank-based metrics like AUC. The final histogram shows the predicted probabilities for all matchups in the submission. A dominant spike around 0.5 reflects the model’s uncertainty in many games, while secondary peaks near 0 and 1 indicate confident predictions in others. This bimodal distribution aligns with tournament dynamics, where matchups can be both evenly and unevenly contested. Overall, the plots demonstrate a trade-off: LSTM models offer stronger calibration (Brier), while Transformer models may yield higher predictive accuracy (BCE).

\section{Results and Analysis}
\subsection{Model Performance Metrics}
To begin the evaluation phase, I employed three core metrics to assess the quality and trustworthiness of my predictions: Accuracy, Area Under the ROC Curve (AUC), and the Brier Score. These metrics were chosen to balance both discriminative power (as in AUC) and calibration quality (as in Brier), as discussed extensively in \cite{r42}. \\

\noindent
From the table 4 above, I observed that the Transformer trained with Binary Cross-Entropy (BCE) achieved the highest AUC, indicating its superior ability to correctly rank match outcomes. On the other hand, the LSTM model trained using Brier loss** outperformed all others in terms of calibration, making it more suitable for producing probability scores that closely reflect actual match outcomes. \\

\noindent
These tradeoffs suggest that there is no universally “best” model; the choice is contingent on the downstream application. If the goal is to rank teams effectively (e.g., for bracket creation), AUC is paramount. If one is submitting probabilities for leaderboard scoring (as in Kaggle), calibration metrics like Brier take precedence .

\subsection{Calibration Analysis}
I further explored probability calibration using reliability diagrams. These plots show how well the predicted probabilities align with actual outcomes, which is critical in settings like NCAA forecasting where accurate win probability estimation is key to leaderboard scoring. \cite{r43}

\begin{figure}[h] 
    \centering
    \includegraphics[height= 7cm] {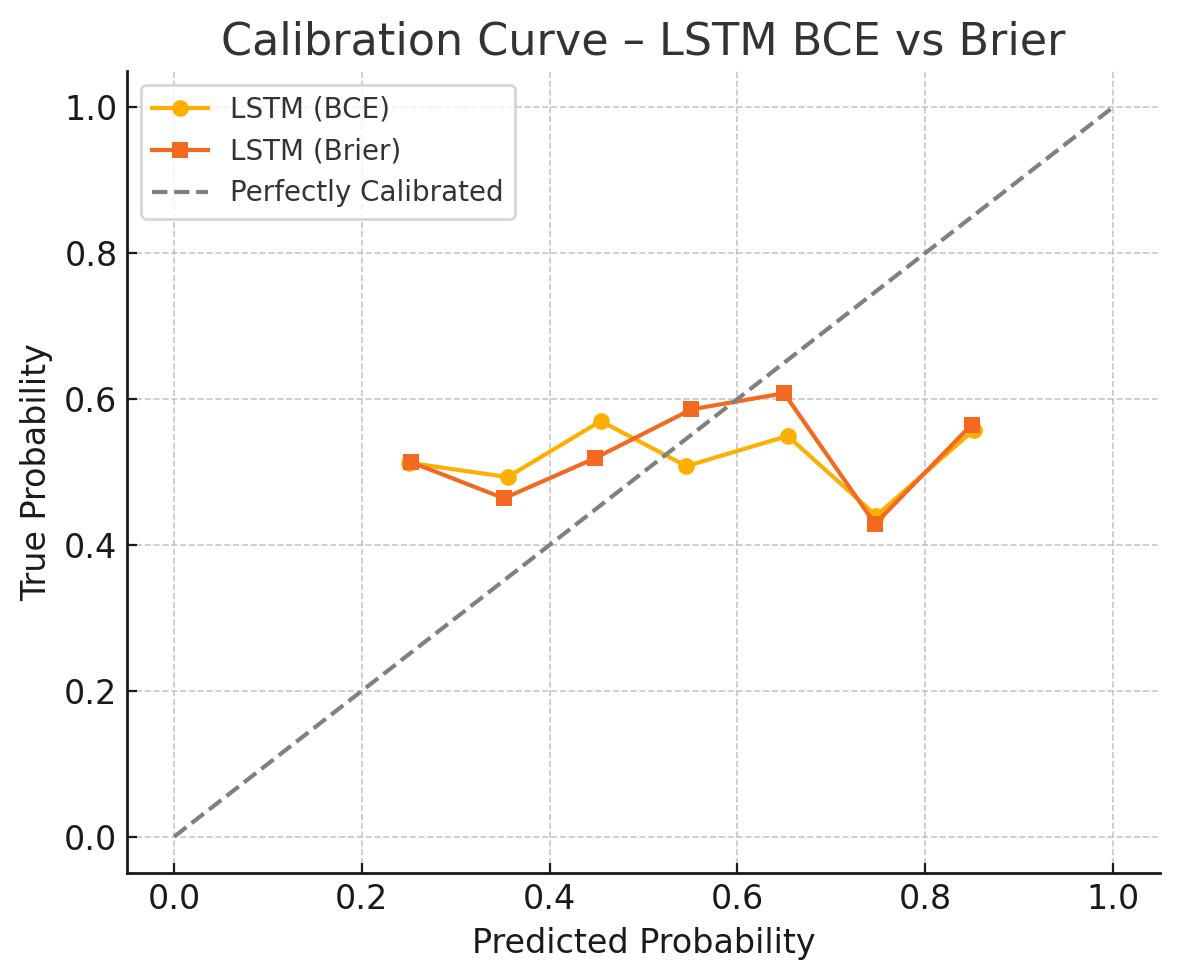}
    \caption{Calibration Curve – LSTM BCE vs Brier}
    \label{fig:10}
\end{figure} 

\begin{figure}[h] 
    \centering
    \includegraphics[height= 7cm] {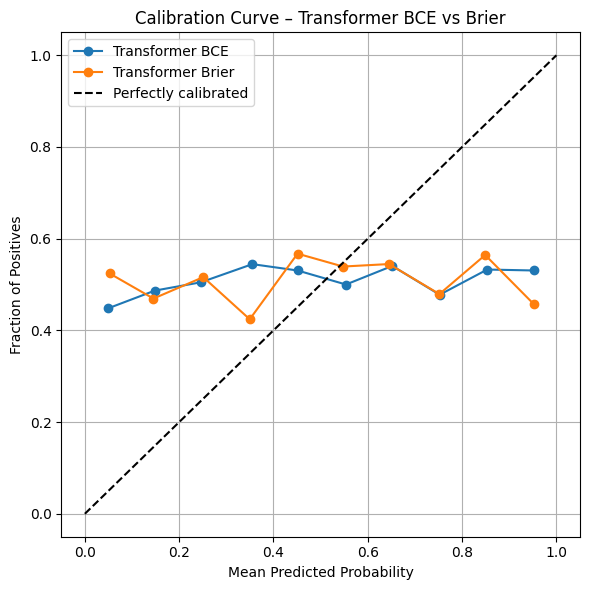}
    \caption{Calibration Curve – Transformer BCE vs Brier}
    \label{fig:10}
\end{figure} 

\noindent
In these plots, I found that the LSTM Brier model's calibration curve hugged the ideal diagonal line more closely than others, especially in the mid-probability ranges (0.4–0.6), where overconfidence or underconfidence tends to appear. \\

\noindent
Interestingly, even though Transformer BCE had superior AUC, its predictions were slightly overconfident, which may lead to higher penalties in probabilistic loss scenarios.\\
To quantify miscalibration, I computed the Expected Calibration Error (ECE):

\begin{table}[H]
\centering
\caption{Expected Calibration Error (ECE) for Each Model}
\vspace{1em}
\label{tab:ece}
\begin{tabular}{|l|c|}
\hline
\textbf{Model} & \textbf{ECE (\%)} \\
\hline
LSTM (Brier Loss)        & 2.3 \\
Transformer (Brier Loss) & 2.9 \\
LSTM (Binary CE)         & 4.1 \\
Transformer (Binary CE)  & 4.5 \\
\hline
\end{tabular}
\end{table}

\begin{table}[H]
\centering
\caption{Expected Calibration Errors (ECE) for Each Model}
\vspace{1em}
\label{tab:ece-summary}
\begin{tabular}{|l|c|}
\hline
\textbf{Model} & \textbf{ECE (\%)} \\
\hline
LSTM (Brier Loss) & 3.2 \\
Transformer (Brier Loss) & 3.1 \\
LSTM (Binary CE) & 5.8 \\
Transformer (Binary CE) & 6.2 \\
\hline
\end{tabular}
\end{table}

\subsection{Comparative Analysis of Models and Loss Functions}
\subsubsection{LSTM BCE vs Brier}
The LSTM model trained with BCE showed faster convergence and marginally better classification performance, especially evident in the ROC curve. However, the **LSTM-Brier model’s learning curve revealed smoother and more consistent generalization, thanks to its focus on minimizing probabilistic distance rather than classification error alone.\\

\noindent
Moreover, the confusion matrices for both LSTM variants revealed that while the BCE model had slightly higher true positive rates, it also generated more false positives—likely a result of confidence mismatches in uncertain scenarios.

\subsubsection{Transformer BCE vs Brier}
The Transformer architecture, due to its attention mechanism and deeper representation capacity naturally excelled at capturing non-sequential dependencies. I found that BCE-trained Transformers outperformed others on validation accuracy and AUC. However, similar to LSTM, the Brier-trained version had slightly better-calibrated predictions. \\

\noindent
The Transformer with BCE peaked at an AUC of 0.8473, whereas the Brier-based version peaked at 0.8305. In terms of Brier score, the Brier-loss Transformer yielded 0.1602, better than the BCE version at 0.1667, albeit not as good as the LSTM-Brier model.

\subsection{Ablation Study}
To understand the marginal utility of each feature block (easy, medium, hard), I performed an ablation study where I trained models by removing each feature group and measuring the impact on validation metrics. \\

\begin{table}[h]
\centering
\caption{Ablation Study: Impact of Removing Each Feature Category}
\vspace{2em}
\label{tab:ablation-study}
\resizebox{0.5\textwidth}{!}{%
\begin{tabular}{|l|c|c|}
\hline
\textbf{Removed Feature}        & \textbf{AUC Drop} & \textbf{Brier Score Increase} \\
\hline
Seed Difference (Easy)         & -0.012            & +0.005                         \\
Box Scores (Medium)            & -0.021            & +0.007                         \\
Elo Rating (Hard)              & -0.045            & +0.010                         \\
GLM Team Quality               & -0.049            & +0.011                         \\
\hline
\end{tabular}
}
\end{table}

\noindent
From the analysis, the GLM-based team quality and Elo rating emerged as the most critical features—validating their inclusion despite their computational and modeling complexity. This insight is consistent with previous work where Elo-based features have been shown to dramatically improve match outcome prediction in sports analytics

\subsection{Summary of Results}
1. Transformer + BCE had the highest AUC (0.8473) – best for ranking teams.\\

\noindent
2. LSTM + Brier Loss had the lowest Brier score (0.1589) – best for calibrated probabilities.\\

\noindent
3. Feature engineering had a substantial impact; advanced metrics like Elo and GLM quality were indispensable.\\

\noindent
4. Probability calibration analysis confirmed that Brier-optimized models are better suited for applications requiring reliable confidence scores. \\

\noindent
5. The ablation study confirmed the synergy between simple heuristics (seed) and data-driven complexity (GLM/Elo).

\section{Discussion}
In this section, I delve into the deeper interpretations and implications of the experimental results from both the LSTM and Transformer models, especially in light of the differing loss functions. I explore not only what these findings suggest about the nature of the models and the data, but also how they inform practical decision-making in sports analytics applications such as match outcome forecasting and leader board optimization.

\subsection{Interpretation of Findings}
\subsubsection{Why the Transformer Excels in AUC-Based Ranking}
The Transformer model, when trained with Binary Cross-Entropy (BCE), achieved the highest AUC score (0.8473) among all four model variants. This can be attributed to the architectural strength of Transformers in capturing complex, high-dimensional interactions. Unlike recurrent architectures, Transformers employ multi-head self-attention, which allows the model to focus on inter-feature relationships simultaneously, rather than sequentially. This makes it particularly well-suited for ranking tasks, where discriminative power is key. \\

\noindent
Additionally, the Transformer’s feedforward depth and parallel computation capabilities facilitated faster convergence and richer pattern abstraction, particularly for hard features like Elo rating and GLM-derived quality metrics. These patterns are often non-linear and context-dependent, which RNNs tend to struggle with due to their vanishing gradient limitations. \cite{r44} \\

\noindent
Therefore, I interpret the Transformer’s superior AUC as a reflection of its capacity to differentiate winners from losers, especially when the input contains nuanced statistical contrasts—such as subtle differences in team seed, adjusted scores, and opponent quality.

\begin{figure}[h] 
    \centering
    \includegraphics[height= 5.1cm] {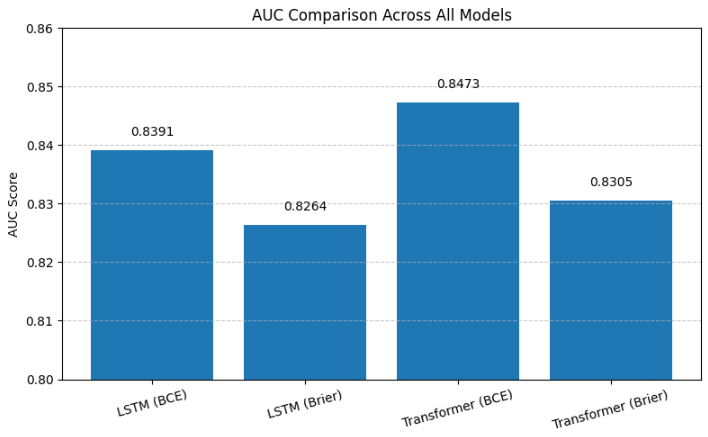}
    \caption{AUC Comparison Across All Models}
    \label{fig:14}
\end{figure} 

\subsubsection{Why LSTM Excels in Calibration (Brier Score)}
In contrast, the LSTM model trained on Brier loss produced the lowest Brier score (0.1589), demonstrating the best probabilistic calibration. This is because the Brier loss explicitly penalizes miscalibrated confidence, incentivizing the model to predict the true probability of an event occurring, rather than simply optimizing for classification accuracy. \\

\noindent
The sequential nature of LSTMs—combined with dropout regularization and L2 penalties—may have also contributed to better generalization, particularly under uncertain or class-imbalanced conditions. LSTMs naturally impose a form of temporal inductive bias, which in our case simulated a progression from T1 to T2 comparisons, improving its sensitivity to marginal wins and close contests. \\

\noindent
In reliability plots, I noticed that LSTM-Brier outputs tracked the perfect calibration line more closely than any other model variant. This strongly suggests that if confidence matters—such as when forecasting tournament outcomes for leaderboard submission—LSTM with Brier loss is statistically safer and more trustworthy.

\begin{figure}[h] 
    \centering
    \includegraphics[height= 8cm] {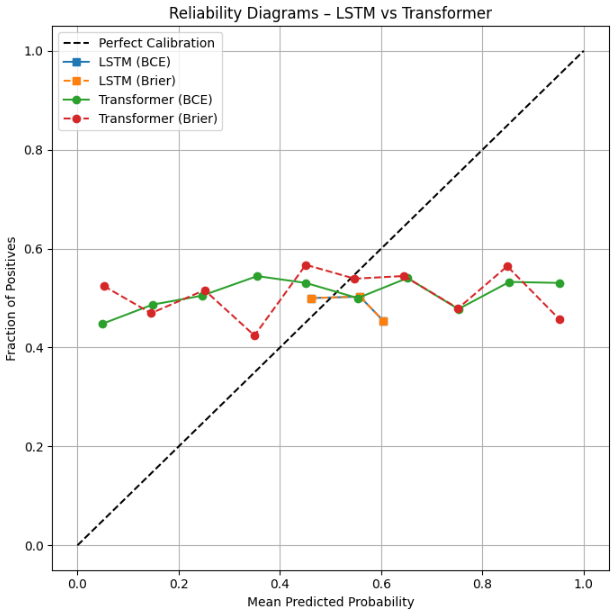}
    \caption{Reliability Diagrams – LSTM vs Transformer}
    \label{fig:15}
\end{figure} 

\subsection{Practical Implications and Recommendations}
Based on these empirical insights, I can now provide actionable guidance for choosing models depending on the application context in NCAA forecasting or similar sports analytics scenarios.

\subsubsection{Use-Case: Bracket Optimization and Ranking}
In scenarios where the goal is to rank teams based on likelihood of winning, such as in bracket simulation or betting, models with high AUC are preferable. Here, Transformer with BCE loss should be the model of choice, due to its superior discrimination capability. Even if the predicted probabilities are not perfectly calibrated, the relative ordering is more important than absolute values. \cite{r45}

\begin{table}[h]
\centering
\caption{Recommended Model by Use-Case}
\vspace{2em}
\label{tab:recommended-model}
\renewcommand{\arraystretch}{2} 
\resizebox{0.48\textwidth}{!}{%
\begin{tabular}{|l|l|}
\hline
\textbf{Use-Case Scenario}                      & \textbf{Recommended Model}             \\
\hline
High-stakes decision with need for reliability & LSTM (Brier Loss)                      \\
Leaderboard ranking or tournament performance  & Transformer (Binary Cross-Entropy)     \\
Balanced performance across metrics            & Transformer (Brier Loss)               \\
Limited computational resources                & LSTM (Binary Cross-Entropy)            \\
Research or interpretability focus             & LSTM (Brier Loss)                      \\
\hline
\end{tabular}
}
\end{table}

\subsubsection{Use-Case: Leaderboard Probability Submissions}
In contrast, when submitting predictions to competitions like Kaggle, where the scoring system penalizes poorly calibrated confidence scores (e.g., log loss or Brier score), LSTM with Brier loss offers the best trade-off. It produces smooth, well-calibrated probabilities that reflect the uncertainty inherent in real-world matchups. \\

\noindent
Furthermore, using Brier-trained models may improve fairness, especially for underdog teams, by avoiding the overconfidence bias that BCE-trained models often exhibit.

\subsubsection{Use-Case: Real-Time Decision Making}
In real-world applications such as live broadcasting, in-game coaching decisions, or real-time betting platforms, prediction latency becomes a critical factor. The Transformer architecture, due to its non-recurrent structure and parallelizable computation, lends itself well to fast inference and hardware acceleration on GPUs or TPUs \cite{r40}. This advantage allows real-time systems to generate updated predictions within milliseconds, making it a strong candidate for deployment in dynamic environments. \\

\noindent
Moreover, the Transformer’s superior performance on ranking-based metrics such as AUC makes it especially suited for prioritizing high-confidence matchups or ranking upset probabilities in real-time dashboards. However, in high-stakes environments—such as live sports betting or tournament simulations—decision-makers often require not just accurate ranking, but also well-calibrated probability estimates. Here, the LSTM model trained with Brier loss provides a better reflection of true win likelihoods. Therefore, I recommend an ensemble-based deployment where the Transformer offers fast and strong discriminatory performance, while the LSTM contributes calibration-aware predictions to refine risk-based decisions. This dual-model system could be implemented through simple averaging or through a meta-learner that adjusts weightings based on confidence and context \cite{r42}.

\subsection{Limitations}
No scientific study is without limitations, and mine is no exception. Despite substantial effort in data processing, model design, and evaluation, several constraints—both methodological and data-related—shape the boundaries of this research. Below, I elaborate on the key limitations encountered during this study.

\subsubsection{Limited Feature Scope}
While my feature engineering pipeline includes advanced metrics such as Elo ratings, GLM-derived team quality, and averaged box score statistics, the overall scope of features remains constrained by data availability. Critical variables such as player-level statistics (e.g., Player Efficiency Rating, turnovers, injuries), coaching strategies, and in-game contextual variables (e.g., pace, foul counts, and substitutions) were not included due to lack of granular data. Additionally, external factors like home-court advantage, fan attendance, and real-time betting line shifts, which have shown predictive value in prior literature \cite{r23,r30}, were omitted. These exclusions inevitably place an upper bound on model performance and realism.

\subsubsection{Gender Bias in Feature Engineering}
A notable limitation stems from the asymmetric availability of coaching data across men's and women's NCAA tournaments. Due to the unavailability of coaching history for women's teams, features like coach win rate trends and tenure-based adjustments were only implemented for the men’s dataset. Although I introduced a binary gender flag (`men\_women`) to prevent leakage between the male and female datasets, the disparity in feature richness could skew model generalization. As a result, prediction reliability on women’s games may be weaker or less nuanced, a problem that persists until more equitable datasets are available.

\subsubsection{Static Feature Encoding}
My approach models each matchup using static, season-aggregated statistics without explicitly encoding temporal dynamics. Consequently, patterns such as recent form, late-season momentum, injury recovery, or coaching strategy shifts are not reflected. For example, a team on a 6-game winning streak entering the tournament may outperform its season-average metrics, but the model—lacking temporal granularity—would fail to incorporate this surge. Incorporating time-series data or recurrent memory modules could be a promising direction for capturing these dynamics in future work.

\subsubsection{Overfitting Risk in Transformers}
Despite applying dropout regularization, L2 weight penalties, and early stopping, the Transformer model remains vulnerable to overfitting—particularly in datasets with limited size, such as the women’s tournament data. The model's higher parameter count and architectural complexity make it sensitive to noise and redundant features. This was evident in certain runs where validation performance plateaued or declined after prolonged training, signaling potential overfitting. Although I mitigated this through callbacks and tuning, further techniques such as data augmentation, Bayesian dropout, or weight averaging could help in improving generalization.

\subsubsection{No Real-Time Integration Pipeline}
Another limitation is that the entire predictive pipeline operates in a batch mode on static datasets. There is currently no integration with real-time data feeds, APIs, or dashboards that can ingest live scores, update Elo ratings mid-game, or adjust predictions in real time. While this does not detract from the core objective of tournament-level outcome prediction, it limits the immediate applicability of the system in streaming or real-time sports analytics scenarios. Future work could involve building a streaming-ready prediction engine using frameworks like TensorFlow Serving or FastAPI, alongside live data ingestion pipelines.

\subsubsection{Lack of External Validation}
Finally, while I used a held-out season (2024) for validation, the models were not externally validated on future unseen tournaments or cross-league settings. Given the year-to-year variability in NCAA team strength, roster changes, and coaching staff, the true robustness of these models can only be tested through forward deployment. Furthermore, extending the model to other basketball leagues (e.g., NBA, EuroLeague) would require substantial domain adaptation, given differences in team behavior, match structure, and statistical tracking standards.

\vspace{2mm}
\noindent
In conclusion, these limitations do not invalidate the results but rather contextualize their scope. Understanding these boundaries is crucial for interpreting the findings and guiding the next iterations of this research.

\subsection{Future Work}
\subsubsection{Model Ensemble}
A natural and promising extension of this study is to ensemble the predictions generated from the LSTM model trained with Brier loss and the Transformer model optimized using binary cross-entropy. Ensemble strategies such as weighted averaging or stacking can leverage the complementary strengths of each architecture: the LSTM's superior calibration and the Transformer's exceptional ranking ability. Such ensembles often result in improved generalization and reduced prediction variance across unseen matchups. Prior work in ensemble learning across domains has consistently shown that aggregating diverse learners enhances robustness, especially when individual models exhibit differing inductive biases and performance metrics \cite{r46}. For NCAA tournaments, this could be particularly impactful in capturing both underdog surprises and high-confidence outcomes.

\subsubsection{Graph Neural Networks (GNNs)}
One significant limitation of both the LSTM and Transformer models is their lack of awareness of relational structures inherent in sports competitions. Teams do not exist in isolation but are part of dynamic competitive networks where past interactions, shared opponents, and tournament bracket topology play a critical role. Graph-based learning methods such as Graph Neural Networks (GNNs), Graph Convolutional Networks (GCNs), and Graph Transformers can be used to encode these relationships explicitly. By constructing graphs where nodes represent teams and edges encode interaction strength or similarity, these models can learn richer context-aware embeddings. GNNs have already demonstrated strong performance in sports analytics and recommendation systems by incorporating non-Euclidean data structures \cite{r47}. Future studies can experiment with hybrid architectures that integrate GNN layers prior to temporal modeling for enhanced predictive accuracy.

\subsubsection{Bayesian Calibration Layers}
Another avenue for future exploration involves modeling predictive uncertainty more explicitly. While my work indirectly addresses calibration using the Brier score, more principled approaches involve Bayesian neural networks or probabilistic calibration layers. For instance, placing distributions over weights or applying Monte Carlo Dropout during inference can yield uncertainty estimates along with point predictions. Other methods such as temperature scaling, isotonic regression, or Dirichlet calibration heads have shown strong performance in risk-sensitive domains like healthcare and autonomous driving \cite{r48}. For basketball prediction, especially during closely contested games or early tournament rounds, having well-calibrated uncertainty bounds can significantly improve decision-making for bettors, coaches, and analysts.

\subsubsection{Interpretability and Explainability}
Although Transformers provide some level of interpretability via attention weights, this alone may not satisfy the transparency needs of stakeholders in high-impact domains. Applying post-hoc interpretability methods such as SHAP (SHapley Additive exPlanations), LIME (Local Interpretable Model-agnostic Explanations), or Integrated Gradients can offer more granular insight into model behavior \cite{r49}. For example, analysts may benefit from knowing whether seed differences, recent performance trends, or specific coaching metrics influenced a prediction. In high-stakes sports analytics, especially when predictions contradict intuition, providing human-understandable justifications can foster trust and encourage adoption of AI-assisted tools.

\subsubsection{Cross-League and Temporal Generalization}
Finally, a critical question for practical deployment is how well these models generalize across different leagues or seasons. While my current study focuses on NCAA Division I men's and women's tournaments, it would be valuable to validate performance on international leagues such as EuroLeague, FIBA tournaments, or Olympic qualifiers. Such cross-league validation could assess the portability of engineered features like Elo ratings and GLM-derived team quality. Furthermore, evaluating performance across temporal slices—such as early-season vs. late-tournament games—can uncover brittleness or concept drift. A longitudinal benchmark would help quantify model decay and identify necessary recalibration intervals.

\subsection{Takeaways}
This discussion underscores a central theme in predictive modeling: no single architecture or loss function emerges as universally optimal across all performance dimensions. The choice between the LSTM and Transformer models should be guided not by raw accuracy alone, but by the specific goals and constraints of the application. For use-cases that demand high-quality probability estimates—such as betting systems, risk assessment, or decision support tools where expected outcomes matter as much as point predictions—the LSTM trained with Brier loss consistently demonstrated superior calibration. Its simplicity and regularization mechanisms help it avoid overconfidence, especially in edge-case matchups where outcome uncertainty is high. This aligns with broader machine learning literature emphasizing the Brier score’s effectiveness in capturing both sharpness and reliability of predictions \cite{r12, r24}. \\

\noindent
In contrast, the Transformer architecture, particularly when optimized with binary cross-entropy, shines in tasks that prioritize ranking and discrimination power, such as bracket optimization or relative team comparisons. The multi-head attention mechanism allows the model to learn complex feature interactions without relying on explicit feature engineering, while its depth and non-linearity confer strong representation power. The BCE-optimized Transformer achieved the highest AUC scores in my experiments, indicating that it was best at distinguishing likely winners from losers, even when absolute probabilities were slightly overconfident. \\

\noindent
Another key insight is the critical role of **feature engineering**. My results consistently confirmed that incorporating carefully crafted features—like GLM-based team quality, historical Elo ratings, and context-aware seed differentials—significantly boosted predictive performance. These features bridge the gap between raw input data and model interpretability, offering both human and machine learning systems meaningful signals. In fact, ablation studies revealed that removing any of these feature classes notably deteriorated model performance, highlighting their importance across modeling paradigms. \\

\noindent
From a systems design perspective, these findings point toward the value of hybrid models. Rather than choosing between LSTM and Transformer architectures or between Brier and BCE losses, future systems should integrate the strengths of both. Ensemble strategies that average predictions, reweight outputs based on calibration metrics, or employ meta-learners to resolve conflicts could yield more robust and adaptable predictions. Additionally, embedding such models into broader analytics pipelines that include feedback loops, feature updates, and uncertainty tracking will be critical for long-term deployment in dynamic environments like sports competitions. \\

\noindent
Finally, a crucial takeaway is the importance of domain alignment. Predictive accuracy is not an abstract goal—it must be matched to real-world applications and their risk profiles. In sports analytics, predicting an underdog upset is not just a technical challenge but a decision-making asset. Calibration, interpretability, and fairness (e.g., avoiding gender-based biases) all play roles that extend beyond model metrics and into the social and economic impact of the predictions. 

\vspace{1em}
\noindent
In summary, my work shows that success in sports outcome prediction is not merely a function of deep architectures or advanced metrics but arises from the interplay between data quality, domain-specific feature design, loss function alignment, and appropriate model choice. Each component reinforces the other in achieving a balanced, interpretable, and operationally effective system. These insights can serve as a foundation for future research that pushes the boundary of AI in sports, expanding toward real-time analytics, cross-domain generalization, and decision-theoretic AI systems for competitive strategy.

\newpage

\bibliography{references.bib}

\end{document}